\documentclass{article}    
\usepackage{hyperref}

\title{Winograd Schemas and Machine Translation}
\author{Ernest Davis \\ 
Department of Computer Science \\
New York University  \\
{\tt davise@cs.nyu.edu}}

\begin{document}
\maketitle

\begin{abstract}
A Winograd schema is a pair of sentences that differ in a single word and
that contain an ambiguous pronoun whose referent is different in the two
sentences and requires the use of commonsense knowledge or world knowledge
to disambiguate.  This paper discusses how Winograd schemas and other 
sentence pairs could be used as challenges for machine translation using 
distinctions between pronouns, such as gender, that appear in the target
language but not in the source.
\end{abstract}

\section{Winograd Schemas}

A Winograd schema (Levesque, Davis, and Morgenstern 2012) is a pair of 
sentences, or of short texts, called the elements of the schema, 
that satisfy the following constraints:

\begin{itemize}
\item[1.]
The two elements are identical, except for a single word or two or 
three consecutive words.

\item[2.]
Each element contains a pronoun. There are at least two noun 
phrases in the element that, grammatically, could be the antecedents of 
this pronoun. However, a human reader will reliably choose one of these 
as plausible and reject the rest as implausible. Thus, for a human 
reader, the resolution of the pronoun in each element is unambiguous.

\item[3.]
The correct resolution of the pronoun is different in the two 
sentences.

\item[4.]
Computationally simple strategies, such as those based on single 
word associations in text corpora or selectional restrictions, will not 
suffice to disambiguate either element. Rather, both disambiguation 
require some amount of world knowledge and of commonsense reasoning.
\end{itemize}

The following is an example of a Winograd schema:

\begin{itemize}
\item[A.]
The trophy doesn't fit in the brown suitcase because it's too 
large.

\item[B.]
The trophy doesn't fit in the brown suitcase because it's too 
small.
\end{itemize}
 
Here, the two sentences differ only in the last word: `large' vs. `small'. 
The ambiguous pronoun is `it'. The two antecedents are `trophy' and `brown 
suitcase'. A human reader will naturally interpret `it' as referring to 
the trophy in the first sentence and to the suitcase in the second 
sentence, using the world knowledge that a small object can fit in a large 
container, but a large object cannot fit in a small container (Davis 
2013).  Condition 4 is satisfied because either a trophy or a suitcase can 
be either large or small, and there is no reason to suppose that there 
would be a meaningful correlation of `small' or `large' with `trophy' or 
`suitcase' in a typical corpus.

We will say that an element of a Winograd schema is ``solved'' if the
referent of the pronoun is identified.

An example of a pair of sentences satisfying conditions 1-3 but not 4 would be

\begin{itemize}
\item[A.]
The women stopped taking the pills because they were pregnant.
\item[B.]
The women stopped taking the pills because they were carcinogenic.
\end{itemize}

Since women cannot be carcinogenic and pills cannot be pregnant, the 
pronoun `they' in these sentences is easily disambiguated using 
selectional restrictions. This pair is therefore not a valid Winograd 
schema.

The Winograd Schema Challenge (WSC)  is a challenge for AI programs. The 
program is presented with a collection of sentences, each of which is one 
element of a Winograd schema, and is required to find the correct referent 
for the ambiguous pronoun. An AI program passes the challenge if its 
success rate is comparable to a human reader. The challenge is administered
by commonsensereasoning.org and sponsored by Nuance Inc. It
was offered
for the first time at IJCAI-2016 (Morgenstern, Davis, and Ortiz, in 
preparation); the organizers plan to continue to offer it roughly once
a year.

\section{Winograd schemas as the basis for challenges 
for machine translation programs}
In many cases, the identification of the referent of the prounoun in
a Winograd schema is critical for finding the correct translation of
that pronoun in a different language. Therefore, Winograd schemas can be used
as a very difficult challenge for the depth of understanding achieved
by a machine translation program.

The third person plural pronoun `they' has no gender in English and
most other languages (unlike the third person singular pronouns
`he', `she', and `it').
However Romance languages such as French, Italian, and Spanish, and Semitic
languages such as Hebrew and Arabic distinguish between the masculine and
the feminine third-person plural pronouns, at least in some grammatical cases.
For instance in French, 
the masculine pronoun is `{\em ils}'; the feminine pronoun
is `{\em elles}'.  In all of these cases, the masculine pronoun is standardly
used for groups of mixed or unknown gender.

In order to correctly translate a sentence in English containing
the word `they' into one of these languages, it is necessary to determine 
whether or not the referent is a group of females. If it is, then the
translation must be the feminine pronoun; otherwise, it must be the 
masculine pronoun. Therefore, if one can create a Winograd schema in
English where the ambiguous
pronoun is `they' and the correct referent for one element is a collection
of men and for the other is a collection of women, then to translate
both elements correctly requires solving the Winograd schema.

As an example, consider the Winograd schema:
\begin{itemize}
\item[A.]
Fred and George knocked on the door of Jane and Susan's apartment, 
but they did not answer.

\item[B.]
Fred and George knocked on the door of Jane and Susan's apartment, 
but they did not get an answer.
\end{itemize}

If these sentences are translated into French, then `they' in the first 
sentence should be translated `{\em elles}', as referring to 
Jane and Susan, and `they' in the second sentence should be translated 
`{\em ils}', 
as referring to Fred and George.

A number of the Winograd schemas already published can be ``converted'' 
quite easily to this form. Indeed, the above example was
constructed in this way; the original form was ``Jane knocked on Susan's 
door, but she did not [answer/get an answer].'' Of the 144 schemas 
in the collection at 
http://www.cs.nyu.edu/faculty/davise/papers/WinogradSchemas/WSCollection.html
there are 33 that can plausibly be translated this way. (\#'s 3, 4, 12, 15, 17, 
18, 23, 25, 26, 27, 42, 43, 44, 45, 46, 57, 58, 
60, 69, 70, 71, 77, 78, 79, 86,  97, 98, 106, 111, 114, 
127, 132, 144) --- essentially any schema that 
involves two people as possible referents where the content of the sentences
makes sense as applied to groups of people rather than individuals.

A similar device, in the opposite direction, relies on the fact that French 
does not distinguish between the possessive pronouns `his' and `her'.
The pronouns `{\em son}' and `{\em sa}' are gendered, but the gender agrees
with the possession, not the possessor. Therefore a Winograd schema that
relies on finding a referent for a possessive pronoun can be turned into
a hard pair of French-to-English translation problems by making one possible 
referent male and the other one female. For example, schema \#124 in the online
collection reads ``The man lifted the boy onto his [bunk bed/shoulders].''
Changing the boy to a girl and translating into French gives
{\em ``L'homme leva la fille sur [ses \'{e}paules/son lit superpos\'{e}]''}. In
the first sentence ``{\em ses}'' is translated ``his'', in the second 
``{\em son}'' is
translated ``her''. The eleven Winograd schemas \#'s 112, 117, 118, 119, 
124, 125,
127, 129, 130, 131, and 143 can be converted this way.

Care must be taken to avoid relying on, or seeming to rely on, 
objectionable stereotypes about men and women. One mechanism that can 
sometimes be used is to include a potentially problematic sentence in 
both directions. For instance, schema 23 from the WSC collection can be 
translated into both ``The girls were bullying the boys so we 
[punished/rescued] them'' and ``The boys were bullying the girls, so we 
[punished/rescued] them,'' thus avoiding any presupposition of whether 
girls are more likely to bully boys or vice versa.

A historical note: Winograd schemas were named after Terry Winograd 
because of a well-known example in his doctoral thesis (Winograd 1970). 
In his thesis, 
this example followed the above form, and the importance of the example 
was justified in terms of machine translation. Winograd's original schema 
was:  ``The city councilmen refused to give the women a permit for a 
demonstration because they [feared/advocated] violence", and Winograd 
explained that, in the sentence with `feared', `they' would refer to the 
councilmen and would be translated as `{\em ils}' in French, whereas in the 
sentence with `advocated', `they' would refer to the women and would be 
translated as `{\em elles}'.  In the later versions of the thesis, published as 
(Winograd 1972), he changed `women' to `demonstrators'; this made the 
disambiguation clearer, but lost the point about translation.

\section{Current state of the art}
No one familiar with the state of 
the art in machine translation technology or the state of the art
of artificial intelligence generally will be surprised to learn that 
currently machine translation program are unable to solve these 
Winograd schema challenge problems. 

What may be more surprising is that
currently (July 2016), machine translation programs are unable to choose
the feminine plural pronoun even when the only possible antecedent 
is a group of women.  For instance, the
sentence ``The girls sang a song and they danced'' is translated into French 
as 
``{\em Les filles ont chant\'{e} une chanson et ils ont dans\'{e}}''
by Google Translate (GT), 
Bing Translate, and Yandex.  In fact, I have been unable to 
construct {\em any\/}
sentence in English that is translated into {\em any\/} language using
the feminine plural pronun.
Note that, since the masculine plural pronoun 
is used for groups of mixed gender in all these languages, it is almost 
certainly more common in text than the feminine plural; hence this 
strategy is reasonable {\em faute de mieux}.  (It also seems likely that the 
erroneous use of a feminine plural for a masculine antecedent sounds even 
more jarring than the reverse.) 

The same thing sometimes occurs in translating the feminine plural pronoun
between languages that have it. GT translates the French word 
`{\em elles}' into Spanish as `{\em ellos}' 
(the masculine form). Curiously, in 
the opposite direction, it gets the right answer; the Spanish `{\em ellas}' 
(fem.)
is translated into French as `{\em elles}'.\footnote{I have heard the
claim made that GT always uses English as an interlingua; that is, when
GT translates from language $S$ to $T$, it always
translates from $S$ to English and then from English to $T$. However,
this example of translating `{\em ellas}' shows that that claim cannot
be entirely true, since the English intermediary could only be `they',
losing the gender.}

\section{Language-specific issues}

The masculine and feminine plural pronouns are distinguished in the 
Romance languages (French, Spanish, Italian, Portuguese etc.)  and in 
Semitic languages (Arabic, Hebrew, etc.) I have consulted with native 
speakers and experts in these languages about the degree to which the 
gender distinction is observed in practice. The experts say that in 
French, Spanish, Italian, and Portuguese, the distinction is very strictly 
observed; the use of a masculine pronoun for a feminine antecedent is 
jarringly wrong to a native or fluent speaker.  
``{\em Les filles ont chant\'{e} une chanson et ils ont dans\'{e}}'' sounds as
wrong to a French speaker as ``The girl sang a song and he danced'' sounds
to an English speaker; in both cases, the hearer will interpret the pronoun
as referrinig to some other persons or person, who is male.
In Hebrew and Arabic, this 
is much less true; in speech, and even, increasingly, in writing, the 
masculine pronoun is often used for a feminine antecedent.

Looking further ahead, it is certainly possible that gender distinctions 
will be abandoned in the Romance languages, or even that English will
have driven all other languages out of existence, sooner than AI systems 
will be able to do pronoun resolution in Winograd schemas; at that 
point, this test will no longer be useful.

In some cases, a translation program can side-step the issue by omitting
the pronoun altogether.
For example, GT translates the above sentence
``The girls sang a song and they danced'' into Spanish as
``{\em Las chicas cantaron una canci\'{o}n y bailaban}'' and into Italian as
``{\em Le ragazze hanno cantato una canzone e ballavano.}''
However, with the more complex sentences of the Winograd Schemas, this
strategy will rarely give a plausible translation for both elements of
the schema.

\section{Other languages, other ambiguities}

Broadly speaking, whenever a target language $T$ requires some distinction
that is optional or non-existent in source language $S$, it is possible
to create a sentence $U$ in $S$ where the missing information is not explicit
but can be inferred from background knowledge. Translating $U$ from $S$
to $T$ thus requires using the background knowledge to resolve the ambiguity,
and will therefore be challenging for automatic machine translation.

A couple of examples:
The word `{\em sie}' 
in German serves as both the formal second person prounoun 
(always
capitalized), the third person feminine singular, and the third person 
plural. Therefore, it can be translated into English as either ``you'', 
``she'', ``it'',
or ``they''; and into French as either {\em `vous', `il', `elle',  
`ils'}, or
{\em `elles'}. (The feminine third-person singular 
German `sie' can be translated
as neuter in English and as masculine in French because the three languages
do not slice up the worlds into genders in the same way.)
Likewise, the possessive pronoun
`{\em ihr}' in all its declensions can mean either `her'
or `their'.
In some cases, the disambiguation can be carried out on purely
syntactic ground; e.g. if `{\em sie}' 
is the subject of a third-person singular
verb, it must mean `she'. However, in many case, the disambiguation
requires a deeper level of understanding. Thus, it should be possible
to construct German Winograd schemas based on the words `{\em sie}' or
`{\em ihr}' that have to be solved in order
to translate them into English.  For example,
\begin{quote}
Marie fand f\"{u}nf verlassene K\"{a}tzchen im Keller. 
Ihre Mutter war gestorben. \\
(Marie found five abandoned kittens in the cellar. Their mother was dead.) \\
\hspace*{3em} vs. \\
Marie fand f\"{u}nf verlassene K\"{a}tzchen im Keller. 
Ihre Mutter war unzufrieden. \\
(Marie found five abandoned kittens in the cellar. Her mother was displeased.) 
\end{quote}

Another example: French distinguishes between a male friend 
`{\em ami}' and a 
female friend `{\em amie}'. Therefore, in translating the word ``friend'' into
French, it is necessary, if possible, to determine the sex of the friend;
and the clue for that can involve an inference that, as far as AI programs
are concerned, is quite remote. Human readers are awfully good at picking
up on those, of course. In general, with this kind of thing, if you want to
break GT or another translation program, the trick is to place
a large separation between the evidence and
the word being disambiguated. In this case, at the present time, the
separation can be pretty small. GT correctly translates ``my friend Pierre''
and ``my friend Marie'' as ``{\em mon ami Pierre}'' 
and ``{\em mon\footnote{The masculine
pronoun `{\em mon}' is standardly used in this phrase rather than the 
feminine `{\em ma}', purely for reasons of euphony.} 
amie Marie}'' and it translates
``She is my friend'' as ``{\em Elle est mon amie}'', but rather surpringly
it breaks down at ``Marie is my friend,'' which it translates ``{\em Marie est
mon ami}.'' As for something like ``Jacques said to Marie, `You have always
been a true friend,'\,'' that is quite hopeless. GT can surprise one, though,
in both directions; sometimes it misses a very close clue, as in ``Marie
is my friend'', but other times it can carry a clue further than one
would have guessed.

\section*{Acknowledgements}
Thanks to Arianna Bisazza, Gerhard Brewka, Antoine Cerfon, Joseph Davis, Gigi 
Dopico-Black, Nizar Habash, Leora Morgenstern, Oded Regev, Francesca 
Rossi, Vesna Sabljakovic-Fritz, and Manuela Veloso for help with the various
languages and discussion of gendered pronouns.

\section*{References}
\noindent
E.~Davis, ``Qualitative Spatial Reasoning in Interpreting Text and Narrative'' 
{\em Spatial Cognition and Computation,} 13:4, 2013, 264-294.

\noindent
H.~Levesque, E.~Davis, and L.~Morgenstern, ``The Winograd Schema Challenge,''
{\em KR 2012}.

\noindent
L.~Morgenstern, E.~Davis, and C.~Ortiz, ``Report on the Winograd
Schema Challenge, 2016'', in preparation.

\noindent
T.~Winograd, ``Procedures as a Representation for Data in a Computer 
Program for Understanding Natural Language," Ph.D. thesis, Department 
of Mathematics, MIT, August 1970. Published as MIT AITR-235, January 1971. 

\noindent
T.~Winograd, {\em Understanding Natural Language,} Academic Press, 1972. 

\end{document}